\documentclass{article}
\usepackage[utf8]{inputenc}
\usepackage[T1]{fontenc}
\usepackage{amsmath}
\usepackage{amsthm}
\usepackage{amssymb}
\usepackage{graphicx}
\usepackage{xcolor}
\usepackage{hyperref}
\usepackage{todonotes}

\usepackage[square,numbers]{natbib}

\title{Meta-learning in natural and artificial intelligence}
\author{Jane X. Wang}

\begin{document}

\maketitle

\begin{abstract}
Meta-learning, or learning to learn, has gained renewed interest in recent years within the artificial intelligence community. However, meta-learning is incredibly prevalent within nature, has deep roots in cognitive science and psychology, and is currently studied in various forms within neuroscience. The aim of this review is to recast previous lines of research in the study of biological intelligence within the lens of meta-learning, placing these works into a common framework. More recent points of interaction between AI and neuroscience will be discussed, as well as interesting new directions that arise under this perspective.
\end{abstract}

\section{Introduction}

Humans are remarkable for continuously learning throughout the entirety of their lives, from acquiring physical reasoning and language skills at a young age \citep{spelke1992origins,marcus1999rule}, to the ability to reason about the detailed complexities inherent in everyday adult life.
One key quality of this learning is that it happens at multiple scales, both in terms of time and abstraction, in a process termed \textit{meta-learning} or \textit{learning to learn}. The fundamental principle of meta-learning is that learning proceeds faster with more experience, via the acquisition of inductive biases or knowledge that allows for more efficient learning in the future \citep{thrun1998learning,schmidhuber1996simple,schmidhuber1987evolutionary}. 

These favorable properties of meta-learning have recently gained it considerable renewed interest within the deep learning/artificial intelligence community. Despite their tremendous successes in recent years \citep{mnih2015human,silver2016mastering}, deep learning systems still require many orders of magnitude of data than humans \citep{lake2017building,botvinick2019reinforcement}. 
Although early work demonstrated the feasibility for neural networks to discover their own learning rules \citep{bengio1991learning,schmidhuber1993neural}, it was only recently that the field has experienced a resurgence of new research in meta-learning using deep neural networks. This has demonstrated the wide-ranging potential of neural networks to meta-learn all aspects of the learning process. Deep neural networks are typically trained via backpropagation, which adjusts the weights of the neural network so that given a set of input data, the network outputs match some desired target outputs (e.g., classification labels). 
Popular meta-learning techniques have therefore spanned everything from methods for meta-learning the initial weights of the network \citep{finn2017model}, the weight update rule itself \citep{ravi2016optimization,andrychowicz2016learning}, or some nonparametric representation of the inputs that is easier to classify \citep{vinyals2016matching,snell2017prototypical}; to deriving an implicit learning algorithm from a black-box recurrent neural network \citep{wang2016learning,duan2016rl,santoro2016meta} (see \citep{vanschoren2018meta} for a comprehensive review).

On the other hand, the idea of "learning to learn" originated within the psychological sciences many decades prior \citep{harlow1949formation}, and focused on one or few-shot learning of learning sets and educational theory \citep{brown1988preschool}. Given the rapid pace of progress, it's illustrative to examine how different lines of work in psychology, cognitive science, and neuroscience fit within the meta-learning perspective as understood currently in artificial intelligence (AI).
This review aims to demonstrate that meta-learning is prevalent in nature, being naturally multi-scaled, and examines past work centered on the points of interaction between neuroscience and incipient research on meta-learning in the field of artificial intelligence. I then suggest interesting new questions and avenues of research that naturally arise under this framework. 

\section{The scales of meta-learning: across and within lifetimes}

Biological learning, at its fundamental level, is the ability of an organism to represent and adapt to changes and challenges presented to it by the external environment. This adaptation is typically in pursuit of a specific drive or goal, such as survival or reproduction. The challenges that one can face in everyday life are widely varying in scope and duration. Accordingly, there exists a range of learning mechanisms that span these different timescales. 

\begin{figure}[h!]
     \centering
     \includegraphics[width=0.95\linewidth]{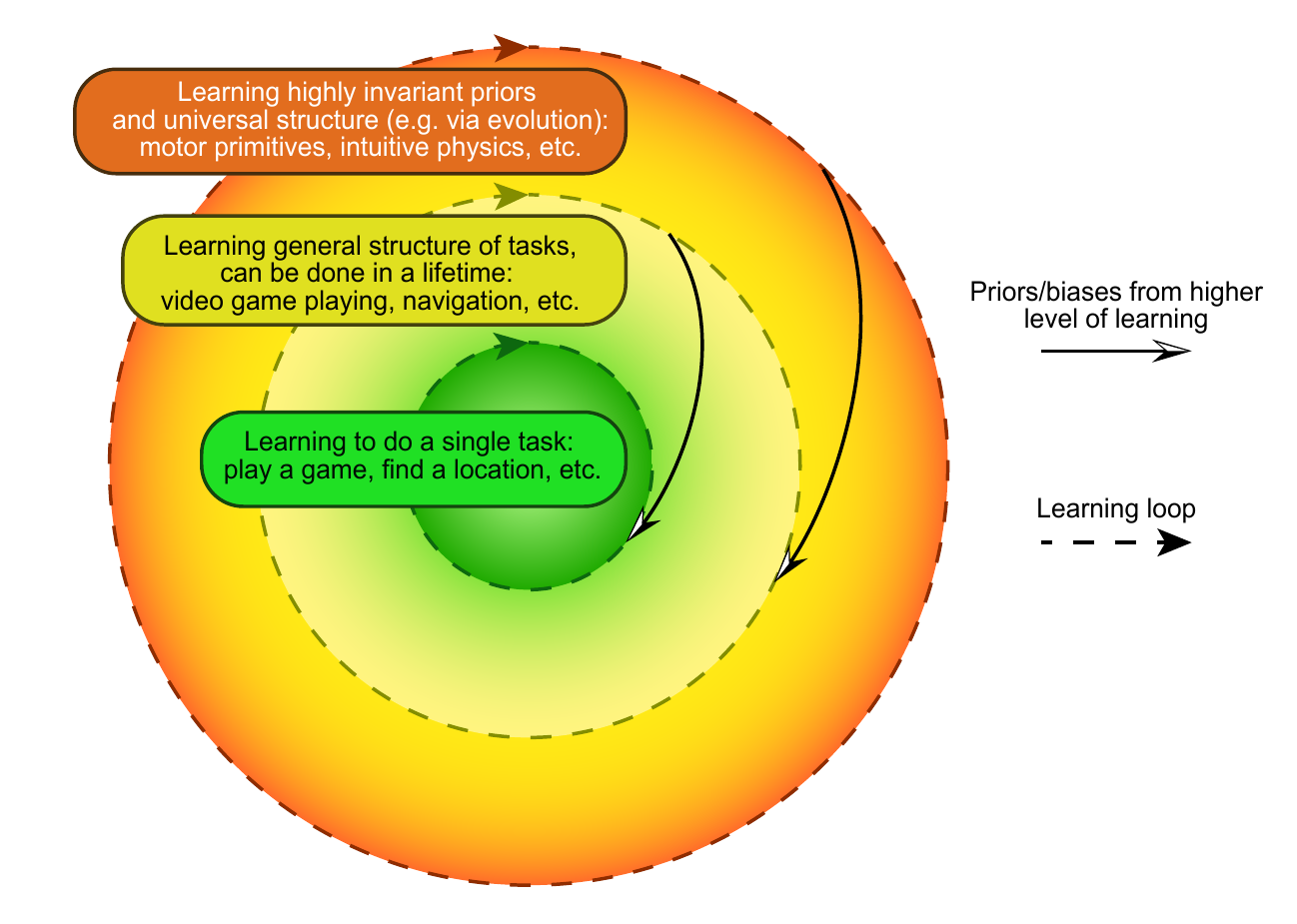}
     \caption{Multiple nested scales of learning in nature. At the highest level, learning is done across generations, via evolution, to learn highly invariant universal structure such as intuitive physics, motor primitives, or other kinds of "core knowledge" \citep{spelke2007core}. These priors help to make learning faster at the level below, where learning is done within a lifetime and involves learning the general structure of different tasks, such as video game playing, how to navigate around a city, or acquiring specific skills. Learning at the innermost level involves fast adaptation within a specific task, such as playing a new video game or finding a certain restaurant within a new city. Again, such fast adaptation is crucially dependent on having learned useful priors and inductive biases at the level above.}\label{fig:fig1}
\end{figure}

There are not only different scales of learning, they are also often nested, such that learning occurring at a longer timescale drives more efficient learning at shorter timescales (see Fig. \ref{fig:fig1}). One of the most interesting examples of this is known as the Baldwin effect \citep{baldwin1896anew}, whereby phenotypic expression of fast adaption and learning creates positive selection pressure, allowing for indirect selection of the genetic basis for these traits to be passed on to future generations.
That faster learning can be selected for by evolution was compellingly demonstrated in simulation by Hinton and Nowlan \citep{hinton1987learning}\footnote{Interestingly, one of the most popular meta-learning approaches, Model-Agnostic Meta-Learning \citep{finn2017model} has been proposed to be closely related to the Baldwin effect \citep{fernando2018meta}.}. In this way, innate (evolutionarily pre-programmed) or developmentally predetermined behaviors interact with learned behaviors and representations \citep{zador2019critique}. For example, the propensity to form place cells (or neurons that tend to fire when only in one particular place in an environment) is innate, while the specific content of these spatial representations in any given environment is learned. The ability to form place cells (and closely related grid cells) thus presumably arose from the benefits conferred by flexibly and quickly representing one's spatial location, which allowed for the evolutionary selection of this innate process. Indeed, the ability to organize and scaffold new knowledge via spatial and relational configurations has been found to be useful for learning even nonspatial conceptual representations in humans \citep{behrens2018cognitive, constantinescu2016organizing}.

Innate learning does not have to be present from birth, but rather can be expressed in relatively stereotyped and predictable trajectories throughout early development. 
According to Alison Gopnik's theory theory \citep{gopnik1999scientist}, human children tend to formulate increasingly complex theories and ways of testing their hypotheses in a relatively predetermined way. Building on this, Elizabeth Spelke posited a core body of knowledge (i.e. object representation, agency, etc.) upon which all other understanding is built, which is present from very early life \citep{spelke2007core} (see also \citep{lake2017building} for a comprehensive review on these topics). That such core knowledge and set developmental trajectories are so conserved indicates their value in building foundational knowledge and skills vital for higher order cognition in humans.

Within a single lifetime, we can see evidence of meta-learning in various animal paradigms of cognition. In one of the first experimental studies of learning to learn, monkeys were challenged to learn an abstract rule for object-role bindings \citep{harlow1949formation}. Two new objects were presented every six trials, only one of which was rewarding, irrespective of object placement. The optimal policy was to choose randomly on the first trial, and then thereafter choose based on the reward outcome of that trial, i.e. perform one-shot learning. Monkeys were able to learn this policy only after an extended period of learning and many sets of new objects. 

Humans tend to meta-learn to much greater extent and at greater levels of abstraction and nesting. For instance, we can perform meta-cognition in order to monitor and improve our own learning progress \citep{metcalfe1994metacognition}, as well as meta-reasoning to perform decision-making given finite computational resources and time \citep{griffiths2019doing}. Learning to learn also has roots within educational psychology and theories of classroom learning and how children learn \citep{bransford2000people,gopnik1999scientist}.
Within cognitive science, hierarchical Bayesian models of cognition capture how learning can occur at multiple scales and via the acquisition of useful, structured priors \citep{gershman2010learning,lake2015human}. This closely parallels the general formulation of meta-learning, and in fact constitutes an exact equivalence for certain forms of meta-learning in AI \citep{grant2018recasting}.

\section{Neuroscience of meta-learning}

While there has been a robust history of meta-learning within the psychological and cognitive sciences, the ties between meta-learning and neuroscience are relatively newer. In this section, I detail several lines of research with direct relevance to meta-learning, and draw ties to corresponding work in AI.

\subsection{Meta-learning as learning of meta-parameters}

Perhaps one of the most straightforward implementations of meta-learning is to learn the parameters of the learning algorithm itself (for instance, the learning rate or discount factor; also called "hyper-" or "meta-parameters").
A notable early account of biological meta-learning proposed that various neuromodulators such as dopamine, serotonin, and noradrenaline played critical roles in the regulation of the meta-parameters of reinforcement learning \citep{doya2002metalearning,schweighofer2003meta}. Relatedly, activity within anterior cingulate cortex (ACC) has been shown to track recent volatility and uncertainty to drive learning rate changes in a Bayesian manner
\citep{behrens2007learning}, and ACC further was proposed to play a central role in dynamically regulating the trade-off between exploration and exploitation during reward-based task learning \citep{khamassi2013medial}. The ACC and certain areas of prefrontal cortex (PFC) have also been suggested to function as a meta-controller dynamically arbitrating between model-free and model-based learning systems \citep{lee2014neural} (see also \citep{daw2005uncertainty}).

Learning of meta-parameters has already been popularized within machine learning, due to its practical benefits on performance (e.g. \citep{jaderberg2017population,xu2018meta}), and lack of needing to hand-tune. Indeed, the momentum has increasingly shifted toward meta-learning more and more aspects of the learning process in recent years \citep{zahavy2020self}.

\subsection{Meta-learning over representations}

Some research lines within neuroscience broadly related to meta-learning are those of learning control over existing representations. In particular, mental schemas \citep{tse2007schemas}
are described as structured mental representations that allow for faster learning, by aiding in retrieval of existing knowledge and integrating new knowledge (see also \citep{van2012schema} for a good review). Such processes are suggested to be mediated by hippocampal-cortical interactions and a specific time course of memory consolidation \citep{tse2007schemas}. In general, the focus of these works center on how existing mental schemas affect new learning, rather than the learning process giving rise to the schema in the first place, and thus can be subsumed within the broader scope of meta-learning.
Another relevant line of research is that on hierarchical representation and cognitive control (the ability to perform task-relevant processing without external support or in the face of distractors) \citep{koechlin2007information}, and a particularly compelling set of works proposing hierarchical organization of prefrontal cortex along the rostro-caudal axis to support increasingly abstract levels of cognitive control \citep{koechlin2003architecture,badre2008cognitive,badre2010frontal}. Such a hierarchically structured organization is intriguingly suggestive of the multi-scaled nature of meta-learning systems. 

Less explored in this area is how such hierarchical representations emerge in the first place. To examine this, it's helpful to turn to developmental neuroscience, which shows that infants can learn latent structure to construct hierarchical rules \citep{werchan20158} and extract statistical regularity from language \citep{saffran1996statistical}. Human adults also learn new structure, and indeed have a bias toward structure learning \citep{gershman2010learning}, even when not strictly needed \citep{collins2013cognitive}, since such structure affords faster learning and generalization in as-yet unseen situations. 
This work ties together hierarchical structure learning and previous proposed theories of basal ganglia-PFC gating models of working memory \citep{o2006making,rougier2005prefrontal}.
Computational accounts have also been put forth that show hierarchical control can emerge implicitly, as a function of training on task distributions for which the optimal policy assumes this hierarchical form \citep{botvinick2004doing}, essentially allowing for optimal learning efficiency on new problems drawn from this distribution.

\subsection{Meta-learning as latent state and Bayesian inference}

Human representation learning and inference is increasingly characterized through a Bayesian lens, and is seen to be key to fast human learning. Lake and colleagues proposed an influential Bayesian framework of learning probabilistic programs, which is able to model how humans acquire new concepts from just a few examples \citep{lake2015human}. Such a model meta-learns by developing structured, hierarchical priors, in which previous experience with similar tasks induce the formation of concepts that improve learning of new concepts. Humans have also been shown to perform tasks by subdividing and decomposing them into optimal hierarchies, which map to optimal policies that are able to generalize to the distribution of all related tasks \citep{solway2014optimal}.

Parallel to these developments in cognitive science, advancements in AI and deep learning have leveraged the ability to train powerful models on large quantities of structured data to learn these efficient representations and learning mechanisms end-to-end \citep{ravi2016optimization, vinyals2016matching, santoro2016meta, xu2018meta,finn2017model, andrychowicz2016learning, wang2016learning, duan2016rl, mishra2017simple}. In the supervised case, it can be shown that a neural network training on an environment of related tasks is equivalent to performing hierarchical Bayesian inference, i.e. learning an appropriate prior (or hypothesis space $\mathcal{H}$) such that the error for learning on new related tasks is minimized \citep{baxter1998theoretical}. This correspondence has also been demonstrated to hold for a popular meta-learning method, model-agnostic meta-learning \citep{grant2018recasting}, in which what is meta-learned are the initial parameters of the neural network, constituting the learned prior. Similarly, memory-based meta-learning approaches \citep{santoro2016meta, wang2016learning, duan2016rl, mishra2017simple} meta-learn the weights of a recurrent neural network, such that the time-dependent evolution of the activity dynamics effectively tracks sufficient statistics of the current task and performs Bayesian updating in a way close to Bayes optimal (given sufficient training) \citep{ortega2019meta}.

Taking this perspective further, we can see that, at least in the context of reinforcement learning, there is little distinction between the fastest inner scale of learning (which integrates incoming information with a learned, structured prior) and latent state inference for decision-making or cognitive control. 
An account for how this process could occur in the brain was put forth by Nakahara et al \citep{nakahara2012learning}, which posited that dopamine encodes more than just reward prediction error, and actually mediates learning more complex reward structure via a learned internal state representation. Further, \citep{donoso2014foundations} suggested that PFC performs online Bayesian inference combined with hypothesis testing to quickly reason over potential strategies which have been learned and stored in memory. Such proposals make contact with a particular subclass of meta-learning models in AI based on episodic memory, in which memory banks of past experiences are incorporated as part of the meta-learning process \citep{santoro2016meta, ritter2018been, wayne2018unsupervised}.

\section{Bridging between AI and neuroscience: New questions and future directions}

We've seen that the multi-scale nature of learning in nature maps well to the framework of meta-learning, as implemented in AI. These points of connection allow us to define new potentially fruitful avenues of research. At the same time, it is important to note the fundamentally different goals of neuroscience compared to AI research. Animal intelligence is already rife with practically useful mental models. Therefore, the driving force of neuroscience is to \textit{discover what already exists}; that is, the representations already acquired by animals and the mechanisms for control over such representations, strategies, knowledge, and subsequent impacts on new learning.

In contrast, the end goal of AI is to \textit{engineer a learning system} from scratch. Deep neural networks are typically initialized with random weights and possess very weak inductive biases \citep{botvinick2019reinforcement}. Therefore, there has been a recognized need for either directly hand-designing algorithmic/architectural biases or learning inductive biases to improve learning. Although the former has been quite successful in achieving state-of-the-art results in various domains, the vast improvements in available training data in recent years have also led to a renewed interest in inducing models to learn these inductive biases through meta-learning approaches. This has inevitably shifted the engineering problem one layer of abstraction, from how to construct \textit{a model that learns} to how to construct \textit{a model of learning itself}. It is in this aspect that cognitive science and neuroscience are well-positioned to offer unique insights to the AI community.

Recent work has already demonstrated how the two can be fruitfully combined, showing that deep reinforcement learning models can capture meta-learning effects similar to how animals learn and in accordance with previous neural findings \citep{wang2018prefrontal}. In a nice demonstration of the "virtuous circle" between AI and neuroscience put forth by Hassabis et al \citep{hassabis2017neuroscience}, these advances have led to resurgent interest in meta-learning within neuroscience, for instance in extending meta-learning to more biologically plausible spiking networks and forms of weight updating \citep{bellec2018long,bellec2020solution}.
More generally, we are witnessing in the last few years great interest in incorporating deep neural networks as models of biological learning \citep{hasson2020direct, cichy2019deep, richards2019deep, marblestone2016toward, botvinick2020deep, ritter2018episodic}, sensory processing \citep{yamins2016using,kell2018task}, or even simultaneously fitted decison-making behavior and neural activity \citep{dezfouli2018integrated}.

Structure learning \citep{gershman2010learning} and model-building are likely to be increasingly important in artificial agent construction, and it's in this area that neuroscience has the potential to offer even more valuable insights. This highlights the need to focus on discovering the processes of structure learning, rather than the structured representations themselves. Furthermore, this perspective has strong implications for task design, emphasizing recording and measuring neural signals during the training process itself rather than current practices of focusing on already trained animals. Additionally, it points to a need for more precise determination of existing priors and biases that animals already possess and how these priors interact with new learning in everyday, complex settings.

\section{Highlights}

\begin{itemize}
\item Multiple scales of learning (and hence meta-learning) are ubiquitous in nature.
\item Many existing lines of work in neuroscience and cognitive science touch upon different aspects of meta-learning, of which we outline three in particular.
\item The distinct but complementary goals of AI and neuroscience point to new points of possible contact, among which meta-learning is well-positioned.
\end{itemize}

\section{Acknowledgements}
The author would like to thank Matthew Botvinick, Kevin Miller, and Kim Stachenfeld for helpful discussions and feedback, and DeepMind for funding.

\bibliographystyle{neuron}
\bibliography{references}

\end{document}